\documentclass[conference]{IEEEtran}
\IEEEoverridecommandlockouts
\usepackage{cite}
\usepackage{amsmath,amssymb,amsfonts}
\usepackage{algorithmic}
\usepackage{graphicx}
\usepackage{textcomp}
\usepackage{xcolor}
\usepackage{comment}
\usepackage{caption}
\usepackage{makecell} 
\usepackage{multirow}
\usepackage{todonotes}
\usepackage{flushend}
\usepackage{mathtools}

\usepackage{tikz} 
\usepackage{pifont}

\def\BibTeX{{\rm B\kern-.05em{\sc i\kern-.025em b}\kern-.08em
    T\kern-.1667em\lower.7ex\hbox{E}\kern-.125emX}}
    
\DeclarePairedDelimiter\norm{\lVert}{\rVert}

\begin{document}

\title{Generalized Learning Vector Quantization for Classification in Randomized Neural Networks and Hyperdimensional Computing
\thanks{
The work of DK was supported by the European Union's Horizon 2020 Research and Innovation Programme under the Marie Skłodowska-Curie Individual Fellowship Grant Agreement 839179.
The work of BAO, JMR, and DK  was supported in part by the DARPA's VIP (Super-HD Project) and AIE (HyDDENN Project) programs.
The work of BAO and DK was also supported in part by AFOSR FA9550-19-1-0241.
}
}

\author{\IEEEauthorblockN{Cameron Diao}
\IEEEauthorblockA{\textit{Department of}\\ \textit{Computer Science} \\
\textit{Rice University}\\
Houston, USA \\
cameron.diao@rice.edu}
\and
\IEEEauthorblockN{Denis Kleyko}
\IEEEauthorblockA{
\textit{UC Berkeley} \\
Berkeley, USA \\
\textit{Research Institutes of Sweden}\\
Kista, Sweden \\
denkle@berkeley.edu}
\and
\IEEEauthorblockN{{\color{black}Jan~M.~Rabaey}}
\IEEEauthorblockA{
\textit{{\color{black}Berkeley Wireless}} \\
\textit{{\color{black}Research Center}} \\
\textit{{\color{black}UC Berkeley}} \\
{\color{black}Berkeley, USA} \\
{\color{black}jan\_rabaey@berkeley.edu}}
\and
\IEEEauthorblockN{{\color{black}Bruno~A.~Olshausen}}
\IEEEauthorblockA{
\textit{{\color{black}Redwood Center for}} \\
\textit{{\color{black}Theoretical Neuroscience}} \\
\textit{{\color{black}UC Berkeley}} \\
{\color{black}Berkeley, USA} \\
{\color{black}baolshausen@berkeley.edu}}
}

\maketitle

\begin{abstract}
Machine learning algorithms deployed on edge devices must meet certain resource constraints and efficiency requirements. 
Random Vector Functional Link (RVFL) networks are favored for such applications due to their simple design and training efficiency. 
We propose a modified RVFL network that avoids computationally expensive matrix operations during training, thus expanding the network's range of potential applications. 
Our modification replaces the least-squares classifier with the Generalized Learning Vector Quantization (GLVQ) classifier, which only employs simple vector and distance calculations. 
The GLVQ classifier can also be considered an improvement upon certain classification algorithms popularly used in the area of Hyperdimensional Computing.
The proposed approach achieved state-of-the-art accuracy on a collection of datasets from the UCI Machine Learning Repository---higher than previously proposed RVFL networks. 
We further demonstrate that our approach still achieves high accuracy while severely limited in training iterations (using on average only $21$\% of the least-squares classifier computational costs).
\end{abstract}


\section{Introduction}

The applications of machine learning are flourishing.
Currently, we see several trends such as the applications of machine learning techniques to challenging problems~\cite{NatureCNN} and the use of machine learning techniques on resource-constrained devices~\cite{SchwartzAI2020}. 
The latter is known under the names ``edge machine learning''~\cite{YaziciEdge2018} and ``tiny machine learning''~\cite{banbury2020benchmarking} since resource-constrained devices are assumed to be located at the edge of computing infrastructure.

Many of the algorithms used in edge machine learning rely on randomness, especially in neural networks~\cite{Scardapane2017} where some of the network's connections may be set randomly.
This operation can be done either at the beginning~\cite{Scardapane2017} or at the end of the network~\cite{FixClassifier2018}.
The former approach is a research area known as 
Random Vector Functional Link (RVFL) networks~\cite{RVFLorig} or Extreme Learning Machines~\cite{ELM06}.
RVFL networks are usually shallow with one hidden layer. Connections between input and hidden layers are set at random and fixed during the training, while connections between hidden and output layers (readout connections) are trained. 
Under this arrangement, the corresponding optimization problem becomes strictly convex; thus, the solution can be derived in a single analytical step using a method like least-squares.
In addition, RVFL networks enjoy theoretical support and have been proven to act as universal function approximators~\cite{Needell2020}.

To improve their operation on hardware, RVFL networks have recently been modified~\cite{intRVFL2020} to have hidden layers with integer-only activations (intRVFL network) using ideas from Hyperdimensional Computing (HDC)~\cite{Kanerva2009}.
This modification, however, still relies on least-squares, the key step in the training part of the network.
Least-squares involves matrix inversion, an undesirable operation on specialized hardware (e.g., field-programmable gate arrays) because it requires a special circuit. 
Therefore, alternative methods for obtaining the readout connections using only simple operations would be preferred.

To address this issue, we propose to replace least-squares with another well-known machine learning technique -- Learning Vector Quantization (LVQ)~\cite{Kohonen1988}.
We use a collection of classification datasets~\cite{Delgado2014} to demonstrate that generalized LVQ (GLVQ)~\cite{Sato1995}, when using one prototype per class, achieves average classification accuracy comparable to that obtained from the least-squares classifier. Furthermore, GLVQ outperforms the the least-squares classifier when multiple prototypes are allowed.

This work contributes three key findings to the intersection of RVFL networks, HDC, and LVQ in machine learning:

\begin{itemize}
    \item From the RVFL point-of-view, GLVQ is an alternative approach for training the readout connections, which requires only a few simple iterative updates of the weights; 
    \item From the HDC point-of-view, GLVQ is a generalized version of centroids-based classification capable of providing higher accuracy than standard centroids classification, which currently prevails in HDC literature on classification;
    \item From the GLVQ point-of-view, the hidden layer activations of the intRVFL network act as a simple nonlinear transformation of the original features, which allows for higher classification accuracy. 
\end{itemize}

The paper is structured as follows. 
The techniques employed in this work are presented in Section~\ref{sec:methods}. 
The proposed approach is presented in Section~\ref{sec:proposed}.
Section~\ref{sec:data} describes a collection of datasets used in the experiments.
The experimental results are reported in Section~\ref{sec:results}.
Section~\ref{sec:conc} concludes the paper.


\section{Methods}
\label{sec:methods}

\subsection{Random Vector Functional Link network}

RVFL networks\footnote{
Here we call them conventional RVFL networks.
}~\cite{RVFLorig} are a type of feedforward neural network with fixed random connections between the input and hidden layers~\cite{Scardapane2017, suganthan2021origins}. 
The network needs only train readout connections between the hidden and output layers, making the task of optimizing network connections a linear least-squares. The conventional RVFL network usually optimizes the readout weights by solving the regularized least-squares (RLS) problem in one analytic step. However, the RVFL network can also optimize these weights with iterative techniques such as least mean squares (LMS) or the fast iterative shrinkage-thresholding algorithm (FISTA)~\cite{suganthan2021origins}.

More formally, the conventional RVFL network trains the readout matrix  $\mathbf{W}^{\mathrm{out}}$ to minimize the mean squared error between the network's predictions and the ground truth. 
For classification, the ground truth is represented by one-hot encodings of the labels. 
Suppose our training set is comprised of $M$ training samples $(\mathbf{x}, \mathbf{y})$ where $\mathbf{x} \in [K \times 1]$ denotes an input feature vector and $\mathbf{y} \in [L \times 1]$ denotes the ground truth label of $\mathbf{x}$. 
Denote $\mathbf{h} \in [N \times 1]$ as the vector of activation values obtained from $\mathbf{x}$. 
We collect vectors $\mathbf{h}^{\top}$ for all $M$ training samples together to form an activation matrix $\mathbf{H} \in [M \times N]$. 
We also form matrix $\mathbf{Y} \in [M \times L]$ storing the ground truth labels of the training set.
We then train optimal weights for readout matrix $\mathbf{W}^{\mathrm{out}} \in [L \times N]$ as follows:
\begin{equation}
    \mathbf{W}^{\mathrm{out}\top} = \min_{\beta} \norm{\mathbf{H} \beta - \mathbf{Y}} + \lambda \norm{\beta}^{2},
    \label{eq1}
\end{equation}
\noindent
where $\lambda$ is a parameter denoting the importance of the L2 regularization term. The closed-form solution to (\ref{eq1}) is given by:
\begin{equation}
\mathbf{W}^{\mathrm{out}\top} = (\mathbf{H}^{\top} \mathbf{H} + \lambda \mathbf{I})^{-1} \mathbf{H}^{\top} \mathbf{Y},
\label{eq2}
\end{equation}
\noindent
in primal space, where $\mathbf{I} \in [N \times N]$ denotes the identity matrix. For more discussion on primal and dual solutions to (\ref{eq1}), see~\cite{suganthan2021origins}.
Now to classify a given input sample $\mathbf{x}$, we first compute activations of the hidden layer $\mathbf{h}$ and then compute predictions $\hat{\mathbf{y}}$ issued by the output layer as $\hat{\mathbf{y}} = \mathbf{W}^{\mathrm{out}} \mathbf{h}$.

The computational complexity of the RLS is dominated by the $N \times N$ matrix inversion in (\ref{eq2}), though the other matrix operations are also important factors.
These matrix calculations typically place a high computational burden on hardware and may cause numerical instabilities~\cite{Villena2014}.
Methods for computing $(\mathbf{H}^{\top} \mathbf{H} + \lambda \mathbf{I})^{-1}$ (with the commonly used QR decomposition, for example) can be parallelized using graphical processing unit (GPU) clusters~\cite{Scardapane2017}. 
These implementations, however, often rely on specific hardware resources such as graphics card computations, which restricts the type of resource-constrained devices capable of training RVFL networks.
We aim to design an RVFL network with computing requirements achievable on a variety of hardware architectures.
Therefore, our goal is to use an alternative classifier to the RLS that will not require matrix decompositions or inversions.

\subsection{Hyperdimensional Computing} HDC~\cite{RachkovskijStructures2001, Kanerva2009, FradySDR2020} also known as Vector Symbolic Architectures~\cite{Gayler2003}, is a neuro-inspired form of computing that represents concepts and their meanings as vectors in a high-dimensional space (hypervectors). 
These hypervectors encode and store information~\cite{Frady17,KleykoPerceptron2020} using distributed representations and are often randomly sampled from the underlying space (e.g., binary). 
The space's high dimensionality ensures that any two random hypervectors are nearly orthogonal with extremely high probability. This property allows HDC operations to produce vector associations that model compositional structures such as sets~\cite{KleykoABF2020}, sequences~\cite{Kanerva2009}, hierarchies, predicate relations, or state automata~\cite{OsipovHD_FSA2017, YerxaUCBHD_FSA2018}. 
Please consult~\cite{KleykoComputingParadigm2021} for a general overview of these representations. 
Using these symbolic operations, HDC can solve a variety of learning tasks with comparable performance to conventional machine learning algorithms~\cite{GeClassificationReview2020, KleykoIndustrial2018, RahimiBiosignal2019}.
In addition, hypervectors can be used as inputs to conventional machine learning algorithms~\cite{RachkovskijClassifiers2007, PSI19, BandaragodaTrajectoryTraffic2019, ShridharEnd2End2020, HyperEmbed}.

A recent modification to RVFL networks~\cite{intRVFL2020} called intRVFL uses ideas from HDC to compute hidden layer activation values.
Prior to computing these values, we generate a matrix $\mathbf{W}^{\mathrm{in}}$ comprised of $K$ $N$-dimensional bipolar hypervectors. 
These hypervectors act as unique labels for individual features in $\mathbf{x} \in [K \times 1]$. 
Next, we quantize the input features and represent each feature as an $N$-dimensional bipolar vector using the thermometer code~\cite{Scalarencoding}.
The thermometer codes for all features are collected in matrix $\mathbf{F} \in [K \times N]$. 
We then bind the $K$ hypervectors in $\mathbf{W}^{\mathrm{in}}$ with the $K$ thermometer codes of corresponding features.
The binding operation is realized as the Hadamard product between $\mathbf{F}$ and $\mathbf{W}^{\mathrm{in}}$.
The resultant hypervectors are nearly orthogonal to each other and can be treated as unrelated.
This holds true even if two different features are represented with the same thermometer code.
We compute activation values $\mathbf{h} \in [N \times 1]$ from these hypervectors using the activation function (\ref{eq10}) described in Section~\ref{sec:proposed}.

We may perform supervised classification on hypervectors $\mathbf{h}$ by leveraging the native properties of HDC~\cite{KleykoBrainlike2014, Kleyko2018}. 
The classifier would construct a single centroid hypervector $\mathbf{w}_{i}$ representing class $c_{i}$ (also called a class prototype). Each class vector $c_{i}$ is expected to exhibit significant similarity to its sample vectors $\mathbf{h}_{i}$, the hypervectors in the training set belonging to $c_{i}$. 
In the simplest case, the class prototype is computed out of its sample vectors using some realization of the bundling operation. A standard choice is to compute the hypervector for class $c_{i}$ to be a centroid (or mean vector) of $c_{i}$ as follows:
\begin{equation}
\mathbf{w}_{i} = \frac{\sum_{j = 1}^{B} (\mathbf{h}_{i}^{(j)})}{B},
\label{eq3}
\end{equation}
\noindent 
where $B$ is the number of class $c_{i}$'s samples in the training set and $\mathbf{h}_{i}^{(j)}$ is the hypervector representing sample $j$ of class $c_{i}$.

Classification using the prototype vectors proceeds straightforwardly: we map test sample $\mathbf{x}$ to hypervector $\mathbf{h}$, with the same mapping used during the training. 
We then calculate the similarities between this hypervector and all class prototypes formed during the training. In this paper, we use squared Euclidean distance as the similarity metric. This metric is tightly connected to the dot product, which is usually used in the HDC context. 
The test sample $\mathbf{x}$ is then assigned to the class whose prototype exhibits maximal similarity to the sample's hypervector $\mathbf{h}$.

The HDC classifier described here has been implemented in many prior works~\cite{Rahimi2016Lang, Najafabadi2016, KleykoIndustrial2018, Rahimi2016EMG, Rasanen2014, Yilmaz2015}.
We note that this classifier solely performs computationally simple operations such as vector addition and similarity metric calculations; it avoids the expensive matrix operations that characterize RLS. 
HDC classifiers in general are suited for edge computing environments due to their ability to run on parallel architectures. 
Several FPGA-accelerated and ultra-low-power implementations of HDC have been proposed~\cite{Montagna2018, Salamat2019, Imani2019FPGA, Schmuck2019}. 
Larger energy savings can be further achieved by using 3D nanoscale devices~\cite{Wu2018}.

However, the centroids-based HDC classifier suffers from capacity and accuracy issues. In particular, any one class centroid representing a large number of training samples may fail to capture intra-class variability~\cite{Witoelar2007}. 
In addition, while centroids capture major similarities amongst hypervectors of the same class, they fail to capture hypervector differences between classes~\cite{Yeseong2018}. 
LVQ is an approach to centroids-based classification that resolves both of these issues. 

\subsection{Learning Vector Quantization}

LVQ is a family of classification algorithms that aims at learning prototypes representing class regions. The regions are defined by hyperplanes between prototypes, which ideally approximate Bayesian borders~\cite{Nova2013}.
The original LVQ formulation by Kohonen~\cite{Kohonen1988} was heuristic and prone to slow convergence problems and instabilities. Modern LVQ learning rules such as GLVQ~\cite{Sato1995} minimize explicit cost functions and so achieve faster convergence and larger flexibility.
These performance improvements in LVQ algorithms have led to their widespread application in fields such as image and signal processing, bioinformatics, and mechanics~\cite{Nova2013}.

We will now formally describe a standard LVQ classifier. Consider a training set comprised of $M$ training samples $(\mathbf{x}, \mathbf{y}) \in \mathbb{R}^{N} \times \{1, \ldots, L\}$.
In this paper, LVQ classifiers consist of a set number of prototypes $\mathbf{w}_{i}^{(j)} \in \mathbb{R}^{N}, j = 1, \ldots, P$ for classes $i = 1, \ldots, L$. We denote the set of all prototypes characterizing a particular LVQ classifier as $\mathbf{W}^{\mathrm{out}}$ since it acts as the readout matrix.
LVQ classification is based on the best matching prototype, which is a winner-take-all strategy. 
Like with HDC, learning aims at finding prototypes so that the training samples are mapped to their corresponding class labels. 
In principle, LVQ classifiers achieve better accuracies than centroids-based HDC classifiers since they approximate Bayesian borders and often have provable bounds on their generalization error~\cite{Crammer2002}.

In this paper, we focus on GLVQ~\cite{Sato1995}, which was proposed as an algorithm that solves the problem of slow convergence associated with the original LVQ algorithm. 
GLVQ defines an explicit cost function from which a learning rule is derived via the gradient descent. This cost function minimizes the generalization error and maximizes the hypothesis margin of the classifier. 
We define the GLVQ cost function as follows:
\begin{equation}
E_{GLVQ} = \sum_{i = 1}^{M} g(\mu(\mathbf{x}^{(i)}, \mathbf{W}^{\mathrm{out}})),
\label{eq4}
\end{equation}

\noindent where $g(\cdot)$ is the logistic sigmoid function parameterized by the slope parameter $\beta$; $\mu(\mathbf{x}^{(i)}, \mathbf{W}^{\mathrm{out}})$ is the relative distance difference:
\begin{equation}
\mu(\mathbf{x}, \mathbf{W}^{\mathrm{out}}) = \frac{d^{+} - d^{-}}{d^{+} + d^{-}},
\label{eq5}
\end{equation}
\noindent 
where $d^{+} = d(\mathbf{x}, \mathbf{w}^{+})$ is the distance of $\mathbf{x}$ from its closest prototype $\mathbf{w}^{+}$ having the correct class label, and $d^{-} = d(\mathbf{x}, \mathbf{w}^{-})$ is the distance of $\mathbf{x}$ from its closest prototype $\mathbf{w}^{-}$ having a different class label. $d^{+}(\mathbf{x}) - d^{-}(\mathbf{x})$ denotes the hypothesis margin of the GLVQ classifier. Using the gradient descent to minimize the cost function, we obtain the GLVQ learning rule:
\begin{align}
\mathbf{w}^{+} &\gets \mathbf{w}^{+} + 2 \alpha \frac{\partial g}{\partial \mu} \frac{2d^{-}}{(d^{+} + d^{-})^{2}} (\mathbf{x} - \mathbf{w}^{+}) \label{eq6} \\
\mathbf{w}^{-} &\gets \mathbf{w}^{-} - 2 \alpha \frac{\partial g}{\partial \mu} \frac{2d^{+}}{(d^{+} + d^{-})^{2}} (\mathbf{x} - \mathbf{w}^{-}) \label{eq7}
\end{align}

\noindent where $\alpha \in [0, 1]$ denotes the learning rate of the classifier and $\frac{\partial g}{\partial \mu} = g(\mu(\mathbf{x})) [1 - g(\mu(\mathbf{x})]$. Further description of the GLVQ learning rule can be found in~\cite{Sato1995}. 

We point to the following connection between GLVQ classifiers and HDC classifiers: several HDC classifiers adopt a perceptron learning rule, which is a simplified version of the GLVQ learning rule, to improve accuracies~\cite{Imani2017, Imani2019AdaptHD, Yeseong2018}. The perceptron learning rule~\cite{Rosenblatt1962} can be formulated as follows:
\begin{align}
\mathbf{w}^{+} &\gets \mathbf{w}^{+} + \alpha \mathbf{x};
\label{eq8} \\
\mathbf{w}^{-} &\gets \mathbf{w}^{-} - \alpha \mathbf{x},
\label{eq9}
\end{align}
\noindent 
for misclassified training sample $\mathbf{x}$, where $\mathbf{w}^{+}$ denotes the prototype with the correct class label and $\mathbf{w}^{-}$ denotes the prototype with the misclassified label. The perceptron learning rule updates $\mathbf{w}^{+}$ and $\mathbf{w}^{-}$ by attractive and repulsive forces from $\mathbf{x}$, same as with the GLVQ learning rule. 
We note that the perceptron learning rule is remarkably similar to Kohonen's original LVQ algorithm~\cite{Kohonen1988}. Both rules employ similar heuristics to learn weight vectors characterizing class boundaries. These heuristics are subject to slow convergence and instabilities on certain patterns of data~\cite{Nova2013, Minsky1969}. We also note that the perceptron learning rule does not have the convergence guarantees or generalization error bounds of GLVQ. In particular, the perceptron convergence theorem only applies to linearly separable data.
Therefore, we suggest that GLVQ should be considered as a more advanced but still very simple technique for improving the standard centroids-based classification in HDC.
 
For classification, we assign input sample $\mathbf{x}$ to the class $c_i$ whose prototype $\mathbf{w}_{i}^{(j)}$ is closest to $\mathbf{x}$ in the input space. GLVQ measures distances $d(\mathbf{x}, \mathbf{w})$ using the squared Euclidean metric. In the case of one prototype per class, this classification scheme produces linear decision boundaries in the activation space, the same as the RLS classifier. In principle, additional prototypes provide the classifier more degrees of freedom and allow for piece-wise decision boundaries that better approximate Bayesian borders~\cite{Witoelar2007}. 
At the same time, too many prototypes increases the generalization error of the classifier, and so there exists a non-trivial optimal number of prototypes to be used for any given dataset~\cite{Crammer2002}. 
We might expect that with the optimal number of prototypes, the GLVQ classifier could better approximate optimal decision boundaries than the centroids-based or RLS classifiers.

\subsection{Computational cost of RLS and GLVQ classifiers}

\begin{table*}[tb]
\renewcommand{\arraystretch}{1.0}
\caption{Specification of computational costs for RLS and GLVQ classifiers.
}
\label{tab:comp:cost}
    \begin{center}
    \begin{tabular}{c c  c c c c }\hline
    \multicolumn{2}{c}{\textbf{RLS Straightforward}} & \multicolumn{2}{c}{\textbf{RLS QR decomposition}} & \multicolumn{2}{c}{\textbf{GLVQ}}  \\ \hline
    Step & Flop number &  Step & Flop number &  Step & Flop number \\ \hline 
     $\mathbf{A}=\mathbf{H}^{\top} \mathbf{H} + \lambda \mathbf{I}$ & $N(2TN-N+2)$ &  $\mathbf{Z}=\mathbf{H}^{\top}\mathbf{Y}$ & $LN(2T-1)$ &  Initialization & $TN$ \\ \hline 
     $(\mathbf{A})^{-1}$ & $2N^3/3$ &  \makecell{Factor $\mathbf{H}^{\top}\mathbf{H} + \lambda\mathbf{I}$ \\ into $\mathbf{Q}\mathbf{R}$} & \makecell{$N(2TN - N + 2) + \phantom{}$ \\ $4N^{3} / 3$} &  \makecell{Distance calculations} & $ITPL(3N-1)$ \\ \hline
    $(\mathbf{A})^{-1} \mathbf{H}^{\top} \mathbf{Y}$ & \makecell{$TN(2N-1)+ \phantom{}$\\$LN(2T-1)$} & $\mathbf{Q}\mathbf{R}\mathbf{w}^{(j)}=\mathbf{z}^{(j)}$ & $2LN(N-1)+LN^{2}$ &  \makecell{Prototypes update} & $IT(6N+PL+19)$ \\ \hline
    Total number & \makecell{$N(2N^2/ 3 + 4TN + \phantom{}$ \\$ 2TL -N -T - L + 2)$} & Total number & \makecell{$N(4N^2/ 3 + 2TN + 2TL + \phantom{}$ \\ $3LN - N - 3L + 2)$} &  Total number & \makecell{$IT(3PLN + 6N $\\$\phantom{} + 19) + TN $} \\ \hline
    \end{tabular}
    \end{center}
\end{table*}

As mentioned earlier, a potential advantage of the GLVQ classifier is that it does not involve the inversion or decomposition of a matrix. 
On the other hand, the GLVQ classifier is a technique involving several iterations (denoted as $I$).
Therefore, it is worth taking a look at the computational costs of training RLS and GLVQ classifiers. 
Table~\ref{tab:comp:cost} presents a summary of their computational costs, where $T$ denotes the number of samples in the training dataset. 
The costs are calculated in terms of the number of floating point operations (flops).  
For the RLS classifier, we considered two approaches for deriving $\mathbf{W}^{\mathrm{out}}$.

One approach involves computing $\mathbf{W}^{\mathrm{out}}$ directly from (\ref{eq2}), the closed-form solution to (\ref{eq1}). However, directly computing the matrix inverse $(\mathbf{H}^{\top}\mathbf{H} + \lambda \mathbf{I})^{-1}$ is numerically unstable and ill-advised. Instead, we consider a different approach that uses QR decomposition to factor $(\mathbf{H}^{\top}\mathbf{H} + \lambda \mathbf{I})^{-1}$~\cite{Villena2014}; this step allows us to solve for the individual rows $\mathbf{w}^{(j)}$ of $\mathbf{W}^{\mathrm{out}}$ without computing matrix inverses. Please consult~\cite{Villena2014} for more details on this approach. We note that computations for this approach can be parallelized using GPU clusters, as mentioned in Section~\ref{sec:methods}. We also note that the number of flops associated with QR decomposition varies based on the implementation algorithm, whether that be Gram-Schmidt orthogonalization, Householder reflections, or Givens rotations. 

In the GLVQ classifier, we could think of three computational steps.
First, we initialize the prototypes with their centroids using (\ref{eq3}).\footnote{When calculating flops in  Table~\ref{tab:comp:cost} we assumed that each class is represented by $T/L$ training samples.}
Second, for each training sample, at every iteration, we calculate distances between the sample and all the prototypes using the squared Euclidean metric.\footnote{Strictly speaking, the number of training samples presented to the GLVQ during one iteration should be a hyperparameter (i.e., a batch size).
In the scope of this work, however, we have kept this number to be equal to the size of the whole training set so $T$ was used when presenting the computational cost.  
}
Finally, the calculated distances are used to find $\mathbf{w}^{+}$ and $\mathbf{w}^{-}$ and to update both prototypes according to (\ref{eq6}) and (\ref{eq7}), respectively.

When looking at the total number of computations needed for each classifier, we could see for the RLS approaches that when $T<<N$, the costs are driven by the $N^3$ term while when $T>>N$, the costs are driven by the $TN^2$ term. 
In the case of the GLVQ classifier, the $ITPLN$ term in the distance calculations becomes the main load, which means that the values of $I$ and $P$ in relation to $N$ are going to determine the computational cost of the GLVQ relative to the RLS.
Below, in Section~\ref{sec:results}, we discuss concrete computational costs for the considered datasets.

\section{intRVFL with GLVQ}
\label{sec:proposed}

\begin{figure}[!t]
    \centering
    \includegraphics[width=1.0\columnwidth]{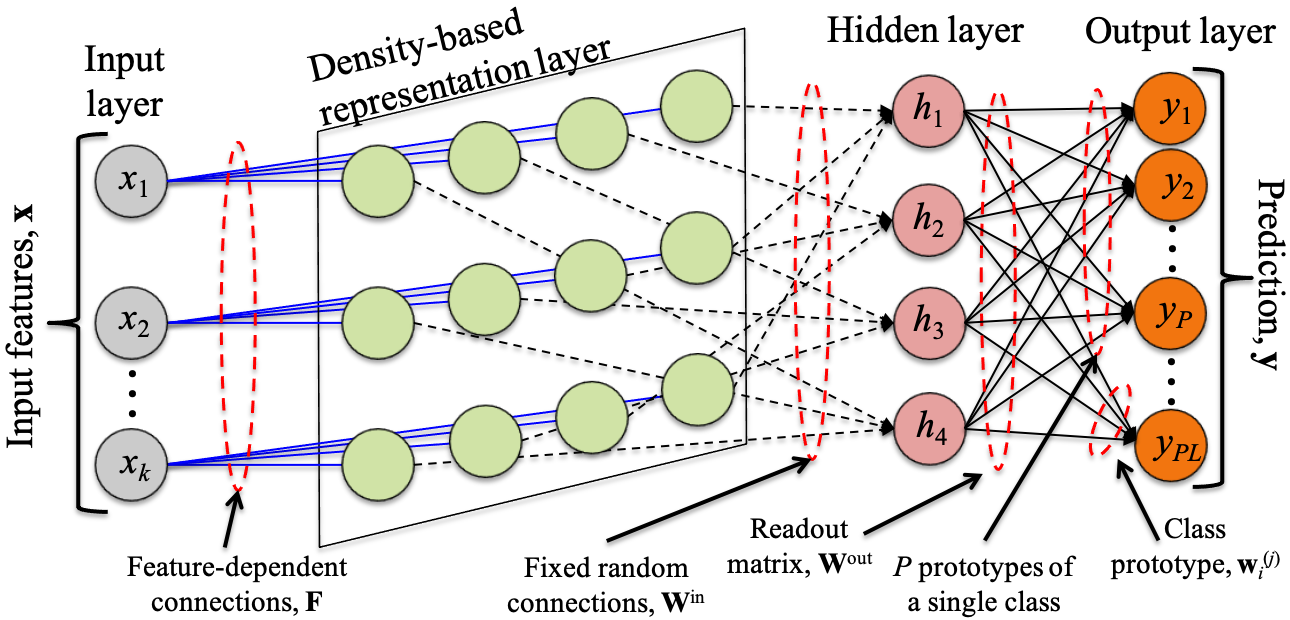}
    \caption{Illustration of the key steps of the proposed approach and concepts involved in its description.
    }
    \label{fig:approach}
\end{figure}

Here we present the approach suggested in this study: where the activations of the hidden layer are formed according to the intRVFL network~\cite{intRVFL2020}, while the readout matrix is formed using the GLVQ classifier~\cite{Sato1995} allowing multiple prototypes per class (see Fig.~\ref{fig:approach}). 
As with the intRVFL, the proposed approach is comprised of four layers: the input layer $\mathbf{x} \in [K \times 1]$ with $K$ neurons, the density-based representation layer $\mathbf{F} \in [K \times N]$ with $KN$ neurons, the hidden layer $\mathbf{h} \in [N \times 1]$ with $N$ neurons, and the output layer $\mathbf{y} \in [LP \times 1]$ with $LP$ neurons; where $L$ denotes the number of classes and $P$ denotes the number of prototypes per class. 
Each neuron $\mathbf{x}_{i}$ in the input layer is connected with $N$ neurons in the corresponding row $\mathbf{F}_{i}$ of the density-based representation layer: $\mathbf{x}_{i}$ is scaled then quantized to the nearest integer as $v = \lfloor \mathbf{x}_{i} N \rceil$, such that $v$ determines the density of active elements in the thermometer code $\mathbf{F}_{i}$ representing $\mathbf{x}_{i}$. Each neuron in the density-based representation layer is connected with only one neuron in the hidden layer. 
Random matrix $\mathbf{W}^{\mathrm{in}} \in \{-1, +1\}^{[K \times N]}$ describes the fixed random connections between the density-based representation and hidden layers. $\mathbf{W}^{\mathrm{in}}$ can be interpreted as $K$ $N$-dimensional bipolar hypervectors. 
We associate the hypervectors of $\mathbf{W}^{\mathrm{in}}$ with the thermometer codes of $\mathbf{F}$ when computing the activation values of the hidden layer. Hidden layer activations can be obtained as follows:
\begin{equation}
\mathbf{h} = f_{\kappa}\left(\sum_{i=1}^{K} \mathbf{F}_i \odot \mathbf{W}^{\mathrm{in}}_i\right),
\label{eq10}
\end{equation}

\noindent
where $\odot$ denotes the Hadamard product used to implement the binding operation; $\sum$ denotes element-wise addition, and $f_{\kappa}$ denotes the clipping function~\cite{intSOM, intESN2020} with $\kappa$ as the threshold parameter.

The proposed approach trains readout connections by training the GLVQ classifier on the hidden layer activation values. 
The readout connections in the GLVQ classifier are a set of prototypes (weight vectors) $\mathbf{w}_{i}^{(j)} \in \mathbb{R}^{N}, j \in \{1, \ldots, P\}$ for classes $i = 1, \ldots L$. Note that assigning variable numbers of prototypes to different classes may affect the generalization error of the classifier~\cite{Witoelar2007}. 
We, however, have not investigated this option in the scope of this work. 

We initialize all prototypes representing class $c_{i}$ as a centroid of $c_{i}$, using (\ref{eq3}). During the training, the GLVQ classifier attempts to learn prototypes that maximize the hypothesis margin; this corresponds to minimizing the GLVQ cost function presented in~(\ref{eq4}). 
In this paper, the cost function employs the sigmoid function as its nonlinear transformation and the squared Euclidean metric as its distance function. Using the gradient descent to minimize the cost function, we obtain the GLVQ learning rule, described in~(\ref{eq6}) and~(\ref{eq7}). 
Gradient descent learning can be realized based on simple stochastic gradient schemes or more advanced approaches like the Broyden-Fletcher-Goldfarb-Shanno algorithm (BFGS). 
The proposed approach employs a limited-memory BFGS (LBFGS)~\cite{Nocedal1999} to realize the gradient descent learning.

Once the GLVQ classifier is trained, we can compute network predictions $\hat{\mathbf{y}}$. 
Given test sample $\mathbf{x}$ with corresponding activation values $\mathbf{h}$, 
we compute the distances of $\mathbf{h}$ from all prototypes $\mathbf{w} \in \mathbf{W}^{\mathrm{out}}$. 
Then $\hat{\mathbf{y}}$ consists of the distances $d(\mathbf{x}, \mathbf{w})$ of $\mathbf{x}$ from all $LP$ prototypes in $\mathbf{W}^{\mathrm{out}}$. We predict $\mathbf{x}$ to belong to the class $c_{i}$ whose corresponding prototype $\mathbf{w}_{i}^{(j)}, j \in \{1, \ldots, P\}$ has the shortest distance to $\mathbf{h}$.

\section{Data, pre-processing, and training}
\label{sec:data}
All comparison tests were performed on a collection of $121$ real-world classification datasets obtained from the UCI Machine Learning Repository~\cite{Dua2019}. 
Here, we used the data and followed the experimental protocol described in the seminal work~\cite{Delgado2014}. 
The only addition we introduced to the collection in~\cite{Delgado2014} was the normalization of input features to the range $[0, 1]$ before passing them through a network. 
There was no additional pre-processing other than that.
For training GLVQ classifiers, we used an LBFGS optimizer with a memory level of $m = 10$.
We trained each classifier for a maximum of $2,500$ iterations or until convergence was reached (convergence as defined by default \texttt{scipy.optimize.minimize} tolerance parameters).
We also initialized prototype vectors to the centroids of training set classes, with additive uniform noise ranging from $(-1, 1)$ in the case of GLVQ. Finally, the reported accuracies in all the experiments were averaged across five independent network initializations.

\section{Experiments}
\label{sec:results}
We wished to evaluate the proposed combination of techniques and our hypothesis that the GLVQ classifier would be a proper replacement for the RLS classifier. Thus, we performed a number of comparison experiments\footnote{The implementations of these experiments are available online at https://github.com/CameronDiao/GLVQ-For-Classification-In-Randomized-NN-And-HDC} on the collection introduced above.  
In the experiments below, we have considered the standard centralized setup where all data was available at once to train a single model, but hypervectors can be also useful in a distributed classification setup~\cite{RosatoHDDistributed2021}.
Five of the experiments compared the GLVQ classifier allowing one prototype per class to the RLS classifier employing L2 regularization (ridge regression). 
We also evaluated the conventional RVFL against intRVFL to confirm findings in~\cite{intRVFL2020}. Then, we evaluated intRVFL against the proposed approach. Finally, we evaluated the KGLVQ classifier employing multiple prototypes per class against the proposed approach.

\subsection{GLVQ vs. RLS: original features}

\begin{figure}[!t]
    \centering
    \includegraphics[width=1.0\columnwidth]{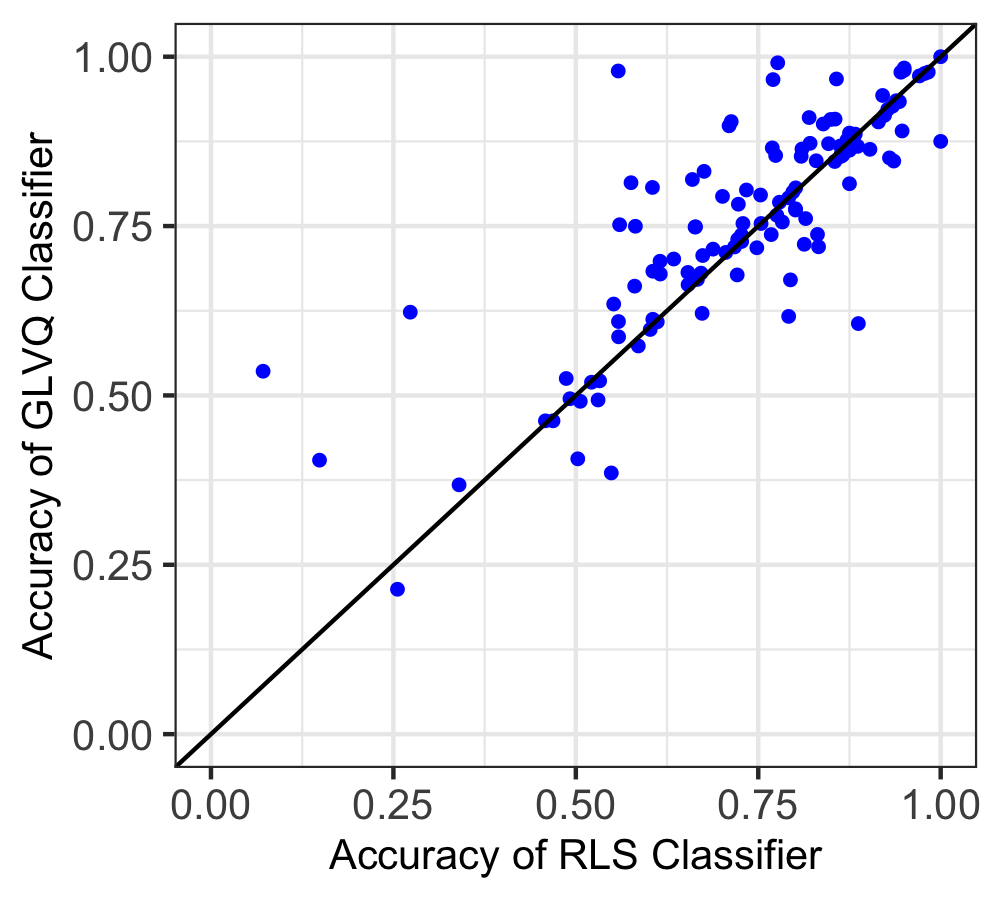}
    \caption{Cross-validation accuracy of the RLS classifier (mean 0.73) against the GLVQ classifier ($P=1$; mean 0.76). 
    }
    \label{fig:test:1}
\end{figure}

In the first experiment, we compared the GLVQ classifier allowing one prototype per class to the RLS classifier. 
Both models accept normalized original feature values as inputs. 
Optimal hyperparameters for each model were obtained using the grid search as described in~\cite{Delgado2014}; $\beta$ for the GLVQ classifier varied in the range $[1, 15]$ with step of size $1$; $\lambda$ for the RLS classifier varied in the range $2^{[-10, 5]}$ with step of size $1$. 
Fig.~\ref{fig:test:1} depicts the results for individual datasets.
We find that the average accuracy of the GLVQ classifier is $0.76$, while the average accuracy of the RLS classifier is $0.73$.

\subsection{GLVQ vs. RLS: RVFL network's hidden layer activations}
\label{sec:GLVQ:RLS:RVFL}

\begin{figure}[t]
    \centering
    \includegraphics[width=1.0\columnwidth]{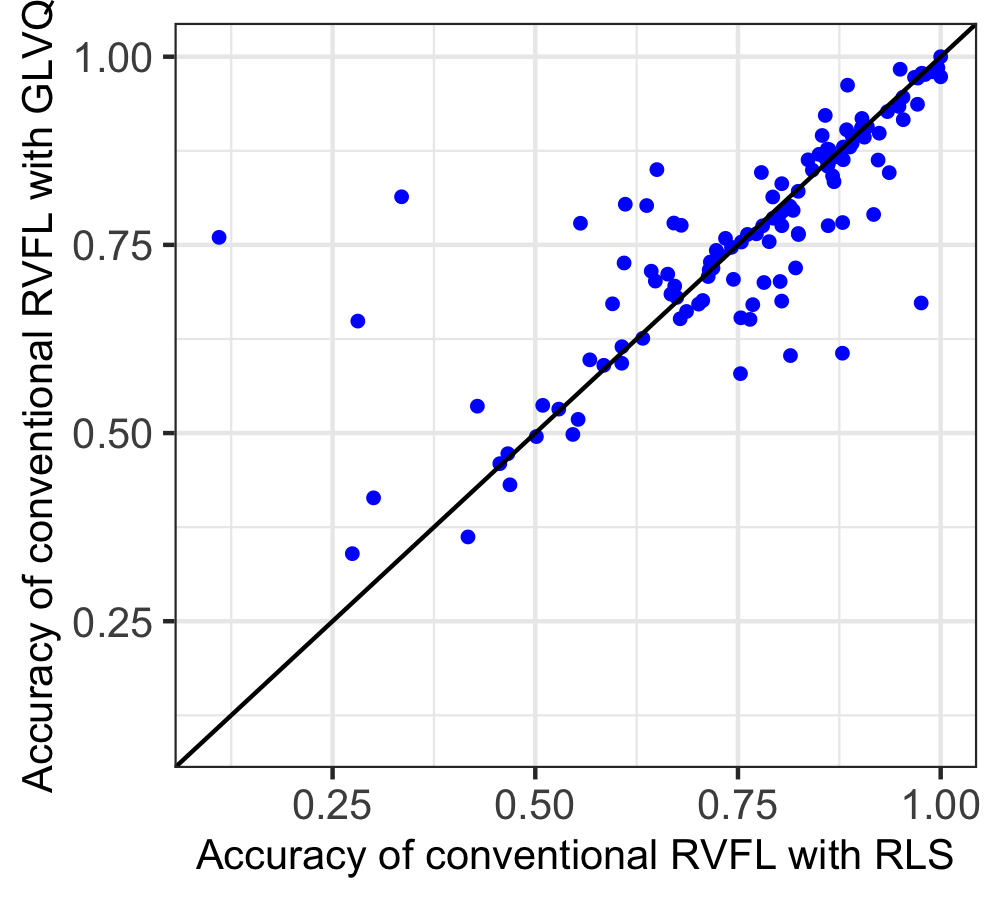}
    \caption{Cross-validation accuracies of two RVFL: RLS classifier (mean $0.76$) vs. GLVQ classifier ($P=1$; mean $0.76$).
    }
    \label{fig:test:2}
\end{figure}

In the second experiment, we compared two variants of the conventional RVFL: the standard one with the RLS classifier and the one with the GLVQ classifier allowing one prototype per class. 
For the RLS classifier, the grid search was used to obtain the optimal hyperparameters of $N$ and $\lambda$. 
For the GLVQ classifier, we used the optimal values of $N$ obtained for the RLS classifier, but we also used the grid search to obtain optimal values of $\beta$ (the same range as in the previous experiment).
Fig.~\ref{fig:test:2} depicts the results for individual datasets.
We see that for some datasets the GLVQ classifier was better, while for others the RLS classifier provided higher accuracy. 
However, importantly, the average accuracies of the variants were both $0.76$. 
We note that the Pearson correlation coefficient between obtained results was quite high, $0.78$.
This result provides us the first indication that the GLVQ classifier can be used in RVFL networks as an alternative way of obtaining the readout connections.

\subsection{Conventional RVFL vs. intRVFL activations}

\begin{figure}[!t]
    \centering
    \includegraphics[width=1.0\columnwidth]{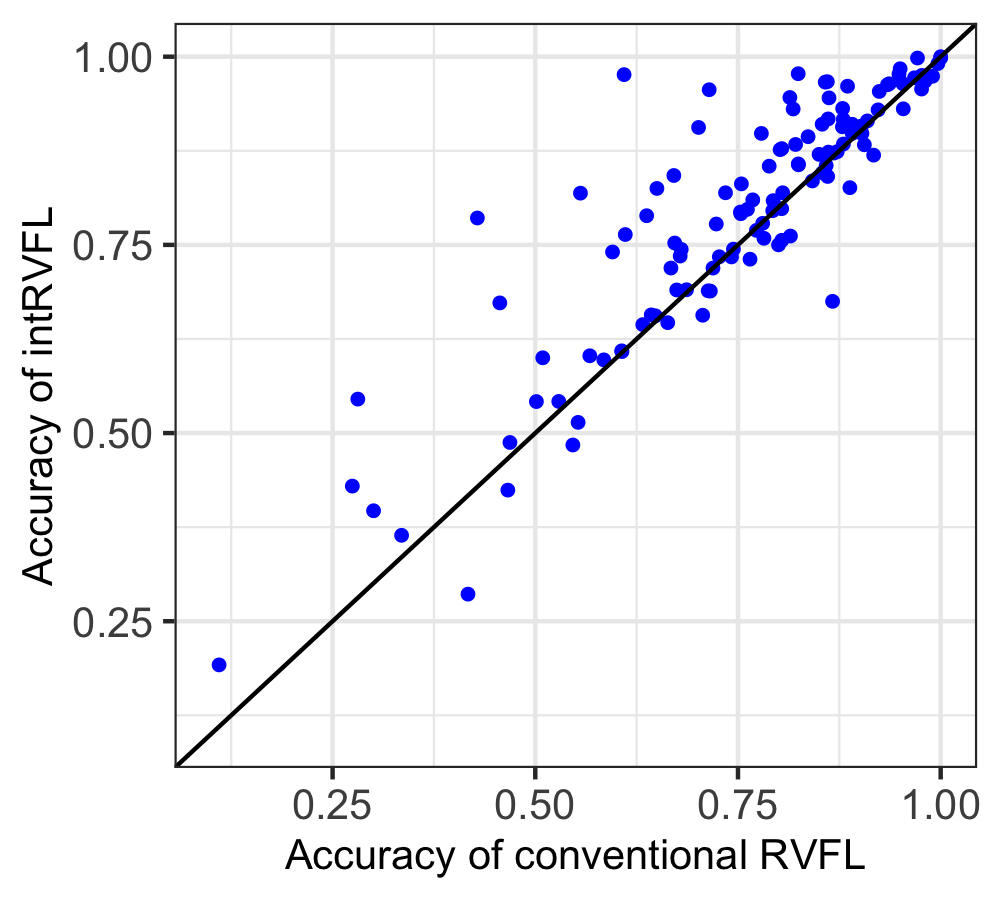}
    \caption{Cross-validation accuracy of the conventional RVFL (mean $0.76$) against the intRVFL (mean $0.80$).
    }
    \label{fig:test:3}
\end{figure}

Next, we compared two different ways of forming the activation values of the hidden layer: the conventional RVFL and the intRVFL. 
For both approaches, the readout connections were obtained using the RLS. 
The optimal hyperparameters for each approach were obtained with the grid search.
This comparison test is meant to confirm findings in~\cite{intRVFL2020}.
Fig.~\ref{fig:test:3} depicts the results for individual datasets.
Note that the results for the conventional RVFL in Fig.~\ref{fig:test:3} are the same as the ones corresponding to the RVFL with RLS in Fig.~\ref{fig:test:2}.
As reported in~\cite{intRVFL2020}, we found that the average accuracy of the conventional RVFL was $0.76$ while the average accuracy of intRVFL was $0.80$.
We also found that the Pearson correlation coefficient between obtained results was $0.88$, similar to the one reported in~\cite{intRVFL2020} ($0.86$).

\subsection{GLVQ vs. RLS: intRVFL's hidden layer activations}

\begin{figure}[t]
    \centering
    \includegraphics[width=1.0\columnwidth]{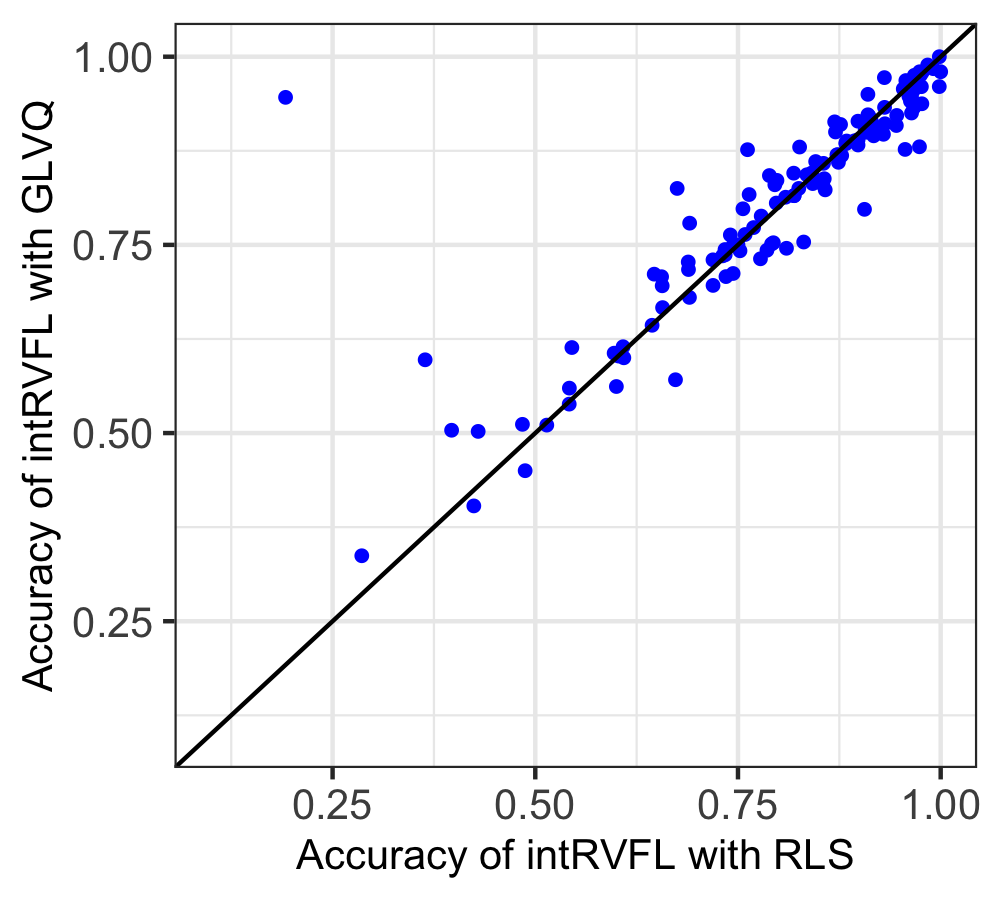}
    \caption{Cross-validation accuracies of two intRVFL:  RLS classifier (mean $0.80$) vs. GLVQ classifier ($P=1$; mean $0.81$).
    }
    \label{fig:test:4}
\end{figure}

Similar to the experiment in Section~\ref{sec:GLVQ:RLS:RVFL}, we compared two variants of the intRVFL: the standard one with the RLS classifier and the one with the GLVQ classifier allowing one prototype per class. 
The search for optimal hyperparameters was done as in Section~\ref{sec:GLVQ:RLS:RVFL}.
Fig.~\ref{fig:test:4} depicts the results for individual datasets.
We found that the average accuracy of the RLS classifier was $0.80$, while the average accuracy of the GLVQ classifier was $0.81$. As in Fig.~\ref{fig:test:2}, there is some difference between the classifiers for individual datasets, but the Pearson correlation coefficient between obtained results remains high (0.87).
This result confirms that the GLVQ classifier was able to leverage the advantage in the intRVFL network's hidden layer activations, since it achieved a performance on par with the RLS classifier and higher than the GLVQ classifier trained on the conventional RVFL network's hidden layer activations.

\subsection{GLVQ vs. RLS: computational costs}

In the previous experiment, we have seen that the GLVQ with $P=1$ performed favorably against the RLS classifier in terms of the average accuracy. 
Recall, however, that in this experiment the maximal number of iterations for the GLVQ was set to a rather large number ($I=2,500$). So when the GLVQ uses maximal iterations, its computational cost would usually be larger than that of the RLS.  
It is, however, worth pointing out that we can limit the GLVQ classifier to a fixed number of iterations during the training. 
The number of iterations might be an important constraint on edge devices because, as we saw in Table~\ref{tab:comp:cost}, it has direct impact on the total computational cost of the GLVQ classifier.
As a demonstration, we modified the previous experiment and limited the GLVQ classifier to only $10$ iterations.
The average accuracy dropped to only 
$0.80$, which still makes it on par with the RLS classifier.
We also calculated the computational costs for each classifier using the results from Table~\ref{tab:comp:cost}. 
For each dataset, the relative costs for the GLVQ and the RLS with QR decomposition were calculated by dividing the costs in flops by the corresponding cost of the straightforward RLS.
The relative costs across datasets were then averaged. 
The average relative cost of the RLS with QR decomposition was $79$\% while that of the GLVQ was only $21$\%.
These results suggest that to conserve resources spent during the training, one could severely restrict the number of iterations for the GLVQ with only a minimal drop in accuracy. 
This flexibility is unavailable when using the RLS classifier.

\subsection{GLVQ allowing multiple prototypes per class}

\begin{figure}[t]
    \centering
    \includegraphics[width=1.0\columnwidth]{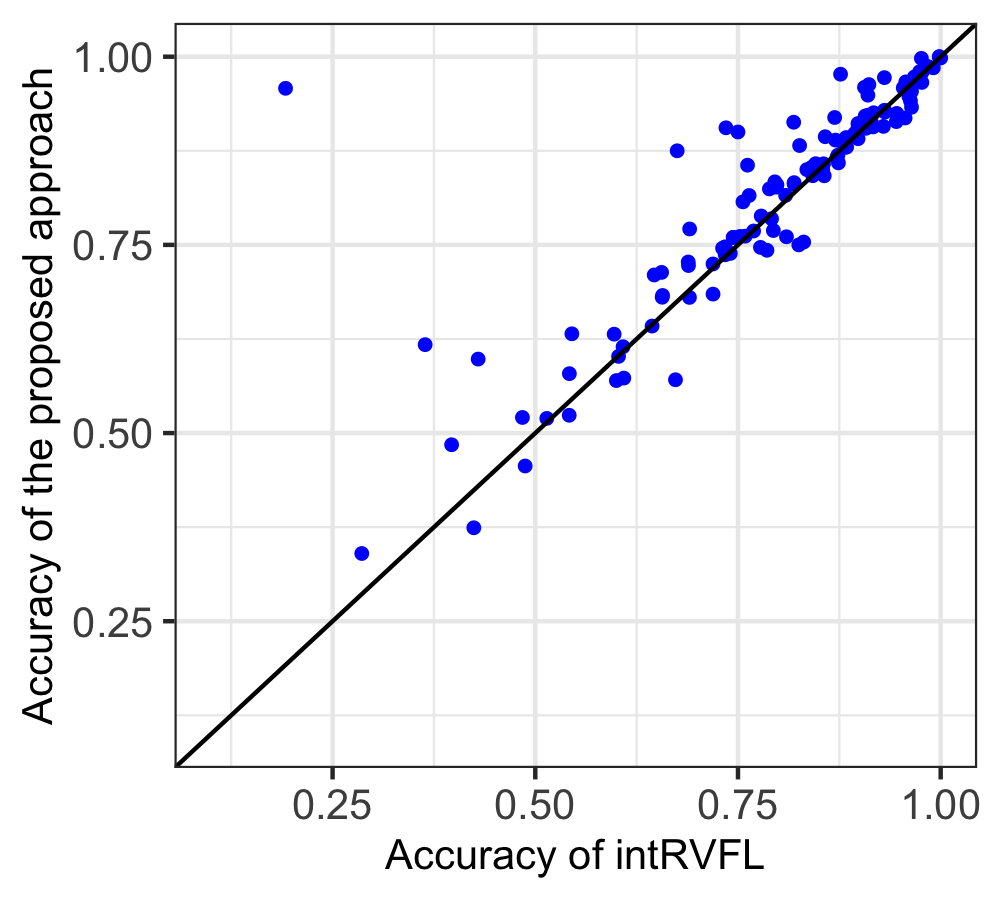}
    \caption{Cross-validation accuracy of intRVFL (mean 0.80) against the proposed approach (mean 0.82).
    }
    \label{fig:test:5}
\end{figure}

In all the previous experiments, the GLVQ classifier was allowed to use only one prototype per class. 
However, as mentioned earlier, the number of prototypes per class is actually a hyperparameter of the classifier. 
In order to assess the performance of the GLVQ at its full potential, we have run the GLVQ classifier where $\beta$ (ranging as before) and the optimal number of prototypes per class (varied between $\{1, 2, 3, 4, 5\}$) were obtained using the grid search. 
The optimal values of $N$ obtained for the RLS classifier were used for a fair comparison. 
Fig.~\ref{fig:test:5} depicts the results of the GLVQ classifier allowing multiple prototypes per class against the RLS classifier for individual datasets.
The average accuracy of the RLS classifier was $0.80$ as before, while the average accuracy of the proposed approach was $0.82$ (Pearson correlation coefficient $0.86$). 
The difference in accuracies was statistically significant by the two-sample paired t-test ($p < 0.05$).
When limiting the GLVQ classifier to only $10$ iterations, the average accuracy of the proposed approach dropped to only $0.81$.

Recall that with $P=1$, the GLVQ classifier is just as accurate as the RLS classifier (cf. Fig.~\ref{fig:test:4}).
Therefore, we conclude that the proposed approach is more accurate than the intRVFL with the RLS classifier because it uses the GLVQ classifier with multiple prototypes per class. 
Multiple prototypes per class allows LVQ classifiers to capture intra-class variability and improve generalization ability~\cite{Witoelar2007}.

\subsection{KGLVQ vs. proposed approach}

We compared the KGLVQ classifier allowing multiple prototypes per class to the proposed approach. The KGLVQ classifier typically uses the Gaussian kernel function, as specified in~\cite{Qin2005}. 
We omitted $9$ UCI datasets from consideration for this comparison: kernel matrix calculations are memory-intensive, and these $9$ datasets had kernel matrices too large for storage in a device with standard RAM capacity. Optimal hyperparameters $N$, $\kappa$ (for intRVFL), $\sigma$ (for the KGLVQ classifier), $\beta$, and the number of prototypes per class (for both classifiers) were obtained using the grid search; $\sigma$ varied in the range $[0.1, 1.1]$ with step of size $0.1$. We obtained this range from the original KGLVQ experiments described in~\cite{Qin2005}.

\begin{figure}[t]
    \centering
    \includegraphics[width=1.0\columnwidth]{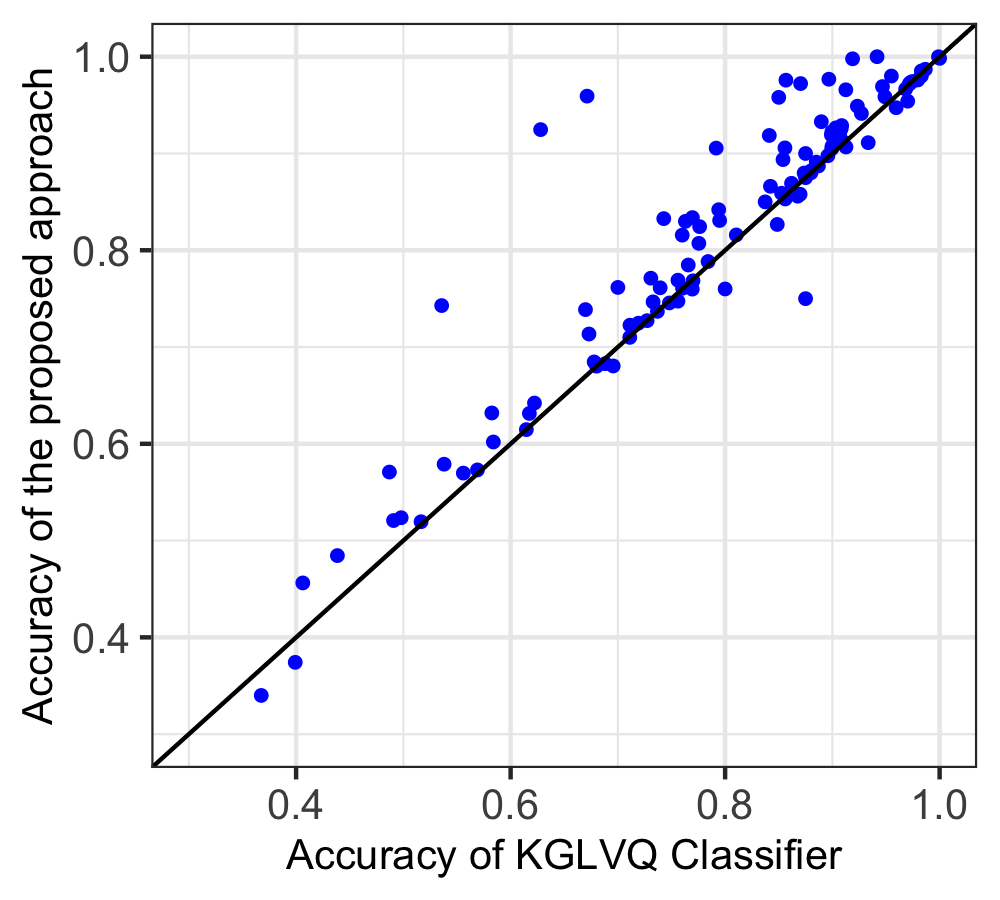}
    \caption{Cross-validation accuracy of the KGLVQ classifier (mean $0.79$) against the proposed approach (mean $0.82$).
    }
    \label{fig:test:6}
\end{figure}

KGLVQ classifiers are infeasible in edge computing environments, since they typically cannot operate on large numbers of data samples concurrently~\cite{Schleif2011}: in addition to kernel matrix calculations, they require the storage and update of large coefficient matrices that implicitly represent prototypes. 
Nevertheless, we are interested in them for their similarities to the proposed approach: both models map inputs to high-dimensional space and adopt the GLVQ learning rule. 
The results suggest that the proposed approach performs better than the KGLVQ classifier. 
On the selected datasets, the average accuracy of the proposed approach was $0.82$  while that of the KGLVQ classifier was $0.79$ (see Fig.~\ref{fig:test:6}). The Pearson correlation coefficient between obtained results was $0.94$.

\section{Conclusion}
\label{sec:conc}

In this paper, we have proposed that the GLVQ classifier replace the RLS classifier in computing the readout matrix of RVFL networks. 
Our experiments demonstrate that the proposed approach achieved higher accuracy than the RLS classifier when the approach used many iterations and assigned multiple prototypes per class.
This setting, however, did not provide computational efficiency over the RLS classifier.
Nevertheless, the efficiency can be achieved by setting the number of prototypes per class to one and severely limiting the number of iterations. For example, with just ten training iterations, the proposed approach achieved an accuracy comparable to the RLS classifier at only 21\% of the RLS's computational cost on average.
Our results suggests that RVFL networks using the proposed GLVQ modification can flexibly improve the accuracy and efficiency of machine learning applications when hardware resources are limited.

\bibliographystyle{IEEEtran} 
\bibliography{references}

\end{document}